\documentclass{article}

\usepackage{PRIMEarxiv}

\usepackage[utf8]{inputenc} % allow utf-8 input
\usepackage[T1]{fontenc}    % use 8-bit T1 fonts
\usepackage{multirow}
\usepackage{booktabs}
\usepackage{makecell}
\usepackage{amssymb,amsmath,amsfonts}
\usepackage{pifont} 
\usepackage{subcaption}
\usepackage{booktabs}
\usepackage{url}
\usepackage[table,xcdraw]{xcolor}
\usepackage{array}
\usepackage{graphicx}

\title{Automatic Prompt Optimization for Knowledge Graph Construction: Insights from an Empirical Study
}

\author{
  Nandana Mihindukulasooriya, Niharika S. D'Souza, Faisal Chowdhury, Horst Samulowitz \\
  IBM Research\\
  \{New York, San Jose\}, USA \\
  \texttt{\{nandana,Niharika.DSouza\}@ibm.com} \\
  \texttt{\{mchowdh,samulowitz\}@us.ibm.com} \\
}

\begin{document}
\maketitle

\begin{abstract}
A knowledge graph (KG) represents a network of entities and illustrates relationships between them. KGs are used for various applications, including semantic search and discovery, reasoning, decision-making, natural language processing, machine learning, and recommendation systems. Automatic KG construction from text is an active research area. 
Triple (subject-relation-object) extraction from text is the fundamental building block of KG construction and has been widely studied, for example, in early benchmarks such as ACE 2002~\footnote{https://www.ldc.upenn.edu/collaborations/past-projects/ace} to more recent ones, such as WebNLG 2020~\footnote{https://synalp.gitlabpages.inria.fr/webnlg-challenge/challenge\_2020}, REBEL and SynthIE.

In recent years, a number of works have explored the use of Large Language Models (LLMs) for KG construction. However, handcrafting reasonable task-specific prompts for LLMs is a labour-intensive exercise and can be brittle due to subtle changes in the LLM models employed. Recent work in NLP tasks (e.g. autonomy generation) uses automatic prompt optimization/engineering to address this challenge by generating optimal or near-optimal task-specific prompts given input-output examples. 

This empirical study explores the application of automatic prompt optimization for the triple extraction task using experimental benchmarking. We evaluate different settings by changing (a) the prompting strategy, (b) the LLM being used for prompt optimization and task execution, (c) the number of canonical relations in the schema (schema complexity), (d) the length and diversity of input text,  (e) the metric used to drive the prompt optimization, and (f) the dataset being used for training and testing. We evaluate three different automatic prompt optimizers, namely, DSPy, APE, and TextGrad and use two different triple extraction datasets, SynthIE and REBEL. Through rigorous empirical evaluation, our main contribution highlights that automatic prompt optimization techniques can generate reasonable prompts similar to humans for triple extraction. In turn, these optimized prompts achieve improved results, particularly with increasing schema complexity and text size.
\end{abstract}

% keywords can be removed

\section{Introduction}

Knowledge graph (KG) construction is a widely studied research area due to the importance and usage of KGs in a wide range of applications. The phrase “knowledge graph” has been used in the literature since at least 1972 ~\cite{hogan2021knowledge} and, to date, there exist several hundred methods for KG constuction \cite{zhong2023comprehensive}.

KGs can be constructed either from structured or semi-structured data using mappings such as RDB2RDF~\cite{sahoo2009survey} or RML~\cite{dimou2014rml} or using unstructured data using various information extraction techniques such as Named Entity Recognition (NER), Relation Extraction (RE), and Open Information Extraction (OIE). In the public domain, crowdsourcing has also been used to build large KGs, such as Wikidata~\cite{vrandevcic2014wikidata}. Nevertheless, crowdsourcing is not a feasible option for most industrial KGs where the knowledge is organization-specific and relevant text corpora are often confidential. 

Recent advances in large language models, which have significantly improved performance on core NLP tasks, have transformed the research area of KG construction from text~\cite{conf/vldb/Khorashadizadeh24,10.1145/3701716.3717822}. Most of the early large language model (LLM)-based methods approach the problem in two key ways. The first is as an extraction of factual and common-sense knowledge using pre-trained language models via prompt-based prediction of masked
objects from partial sentences describing
complete triples \cite{petroni-etal-2019-language}. The second is by fine-tuning an LLM on the autoregressive generation of text for triple extraction \cite{melnyk-etal-2022-knowledge, RossielloCMCG23}. Similarly, in-context learning with decoder-only transformer models is applied for relation extraction~\cite{DBLP:conf/esws/Khorashadizadeh23}. Alternatives such as Re2G \cite{glass-etal-2022-re2g} and GraphRAG~\cite{edge2024local} use LLMs to build Retrieval Augmented Generation (RAG) systems \cite{NEURIPS2020_6b493230}.

\subsection{Prompt Engineering: Challenges and Opportunities}
LLMs typically have a large number of parameters and are (pre)-trained on larger corpora, followed by iterative refinement using Reinforcement Learning from Human Feedback (RLHF)~\cite{stiennon2020learning} or Reasoning-oriented Reinforcement Learning~\cite{guo2025deepseek}. These models are growing in popularity due to their capabilities in comprehending detailed instructions for diverse and complex downstream tasks~\cite{10.5555/3600270.3602070} in a broad sense.

More often than not, the downstream performance of LLMs is heavily coupled with the quality of the prompt used to instruct the model. Studies have shown that large language models (LLMs) are quite sensitive to minor variations in prompt phrasing. For example, the addition, removal, or reordering of just a few tokens can lead to significant differences in task performance~\cite{10.1145/3560815, 10.1145/3689217.3690621}. Generic prompts do not typically produce good responses and the most effective prompts are almost always handcrafted by humans. This makes prompt curation a labour-intensive iterative process involving a substantial amount of manual experimentation. This process of optimizing the prompt language to elicit the best possible performance is referred to as ``prompt engineering". Given the human effort involved in the practice, prompt-engineering techniques are often brittle, non-transferable, and suffer from scalability issues ~\cite{arora2023ask}. Currently, prompt engineering is more of an art than a science, as a delicate balance is required in the design process to ensure clarity and specificity in the prompts, avoid ambiguity, and steer appropriate behaviour to obtain the desired output. More importantly, the human labor involved needs to be repeated whenever the underlying LLM is changed, since previously optimized prompts may no longer yield optimal results. For example, this applies when we switch to a different model, to a different parameter-size variant of the same model, or upgrade to new versions of the LLM trained using different strategies/additional data. This is further complicated when the prompt has to take into account a domain-specificity or a different language (direct translations of prompts often produce poor results~\cite{mondshine-etal-2025-beyond}) 

% In some use case it is easier to write prompts manually, even such case you will need domain experts, people with expertise in prompt engineering, etc. 

\subsection{Automatic Prompt Optimization: Motivation}
Recent advances and superior pre-training strategies have resulted in LLMs that have shown remarkable, often superhuman, capabilities to generate reasonable responses even with limited or no examples. This has inspired a few works to explore the capabilities of an LLM in designing the best possible prompts for specific tasks. This process is referred to as automatic prompt optimization~\cite{ramnath2025systematic} or automatic prompt engineering~\cite{li2025survey}. The core idea is that the LLM is tasked with generating task-specific instructions where the task is presented via output demonstrations (with some given examples). It generates several instruction candidates, either via direct inference or a recursive process driven by scoring metrics. The LLM executes these instructions, and the best instruction improving the scoring metric is retained. Automatic prompt optimization can be applied to any task that is solved by prompting LLMs.

In this paper, we explore three prompt optimization approaches, namely, DSPy, TextGrad, and APE (detailed descriptions in Section \ref{sec:apo}), adapted to triple extraction. We extensively evaluate different settings guiding this task. Specifically, we explore the effect of changing (a) the prompting strategy, (b) the LLM being used for prompt optimization and task-execution, (c) the number of canonical relations allowed in the schema (i.e. the schema complexity), (d) the length and diversity of the input text, (e) the metric driving the optimization, and (f) the dataset being used for training and testing. 

We demonstrate that automatic prompt optimization techniques for triple extraction can generate reasonable prompts akin to human-generated prompts and thus achieve improved results. Specifically, the most significant gains were observed as the text size, context length, and schema complexity increases. To the best of our knowledge, this is the first work that analyses the impact of automatic prompt optimization for triple extraction for KG construction.

% https://cameronrwolfe.substack.com/p/automatic-prompt-optimization

\section{Triple Extraction for Knowledge Graph Construction}

To build a Knowledge Graph from text corpora, a system needs to convert the unstructured natural-language text into a set of elemental triples. Triple extraction involves sub-tasks such as identifying the core entities mentioned in the text (Named Entity Recognition or NER), along with identification/extraction of relations that constitute the facts mentioned in the text (Relation Extraction or RE). These sub-tasks can be pipelined, by performing entity extraction first, and then extracting relations and triples or in a joint manner, where entities, relations, and triples are extracted simultaneously (end-to-end relation extraction). 
In this paper, we focus on end-to-end triple extraction, where the input is a natural language text and the expected output is a set of triples. 

Triple extraction may be either in the form of closed relation extraction (where the set of relations is restricted to a set of known canonical relations) or of the open information extraction form~\cite{etzioni2008open}. Pre-curated canonical relations are useful when the resulting KG needs to be grounded by a pre-defined ontology. On the other hand, open information extraction requires relations to be spontaneously discovered from text and then represented using short descriptive phrases. This requires converting the phrases extracted from text to surface forms that are consistent/resolvable throughout the extraction process. Thus, when a large number of sparse and potentially diverse relations are involved, extensive post-processing may be needed. This includes clustering~\cite{zhang2018knowledge}, canonicalization~\cite{dash-etal-2021-open}, and relation linking~\cite{rossiello2021generative} to ground extracted relations to the canonical relations in the KG, which adds a layer of complexity to the KG-extraction process over the LLM-based processing. Since we are interested in evaluating the potential of LLMs for KG, we focus on closed triple extraction with a set of pre-known canonical relations.

In traditional triple extraction approaches, the set of canonical relations is used during the training phase to select the labels for relation classification or ranking. Invoking LLMs passes knowledge about these canonical relations to the model using either fine-tuning or prompt tuning. Instruction fine-tuning via InstructGPT \cite{ouyang2022training}, Reinforcement Learning from Human Feedback (RLHF) \cite{christiano2017deep,stiennon2020learning} is known to assist LLMs in following a broad range of written instructions exactly while generating responses as Chain-of-Thought reasoning steps. On the other hand, in-context learning \cite{min2022rethinking,xie2021explanation} in conjunction with prompt engineering can teach models to perform new tasks by providing a few demonstrations of input-output pairs at inference. 

\begin{figure}
  \centering
  \includegraphics[width=\linewidth]{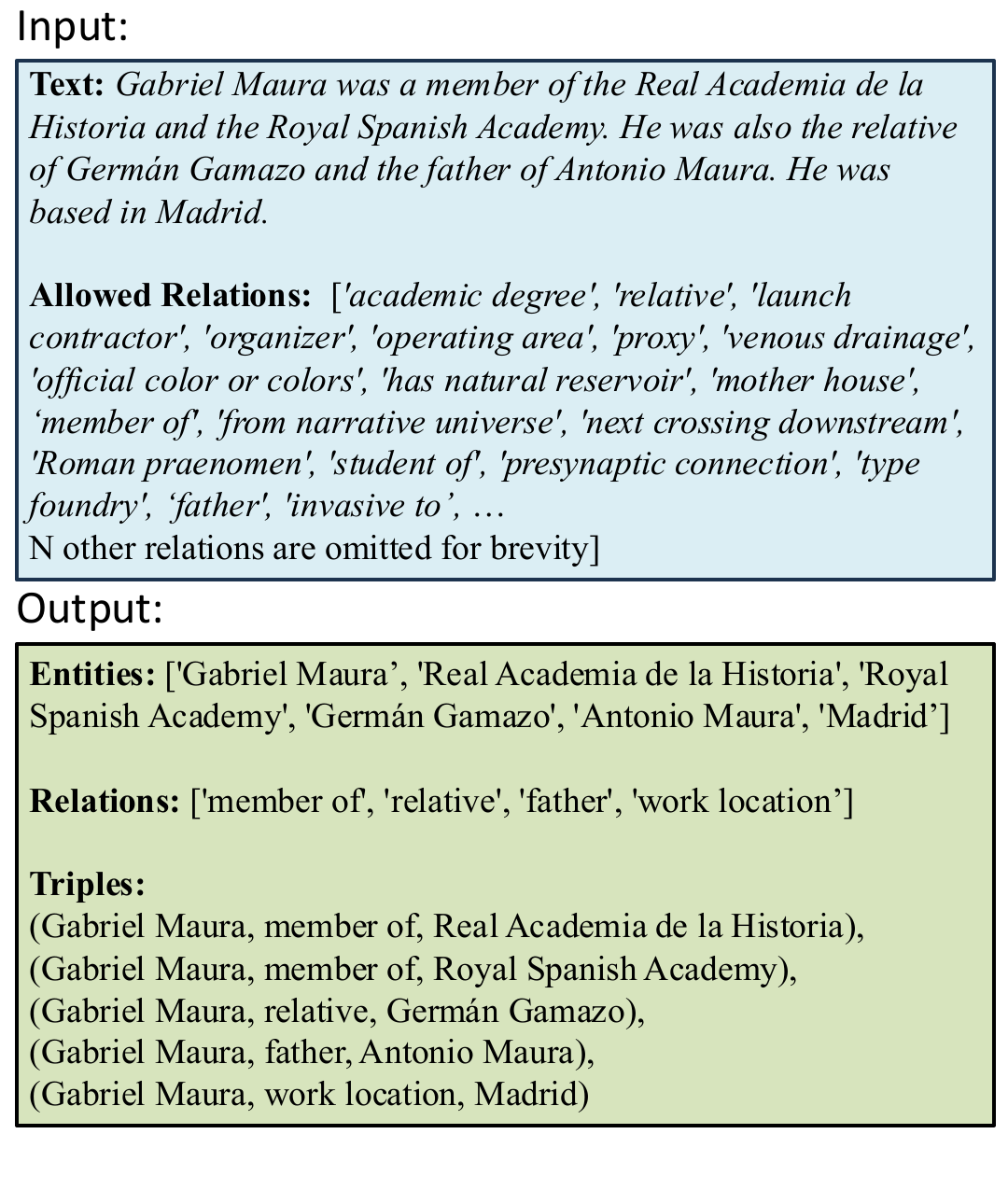}
  \caption{An example of the triple extraction task with inputs and expected outputs (from the SynthIE dataset).}
  \label{fig:task_exp}
\end{figure}

To employ in-context learning or Chain-Of-Thought with a canonical set of relations during prompting, we follow an approach analogous to Text2KGBench~\cite{text2kgbench} and provide the schema information as input in the prompt. Overall, we formulate our task (illustrated in Figure~\ref{fig:task_exp}) as follows: Given an input text written in natural language form and a list of allowed canonical relations, the LLM is asked to extract triples from the text in the form of (subject, relation, object). In addition to the triples, the LLM may be explicitly asked to generate entities and relations as an intermediate output.

\section{Automated Prompt Optimization}
\label{sec:apo}
Recent progress in Natural Language Processing has been exponential due to the availability of powerful LLMs such as GPT-4, Llama, and Claude. Users interact with models by supplying carefully handcrafted prompts, with the task formulated as a token completion problem. As alluded to before, prompt design can be a laborious exercise heavily reliant on domain expertise and ad hoc heuristic evaluation. Automated Prompt Optimization (APO) is a body of work that seeks to address this problem by leveraging systematic data-driven approaches.

The works of~\cite{cui2025automatic} provide a comprehensive taxonomy of automatic prompt optimization frameworks which refine prompts with no or minimal human intervention. Broadly, these methods can be categorized along multiple dimensions. These include:  optimization space (i.e. discrete text-based vs. soft prompting or gradient-based), optimization targets (i.e. instructions vs. examples),  optimization objective (i.e. task performance, safety, or generalizability), operators used to generate new prompts (e.g. purely model-based vs. iterative refinement of example prompts), and iterative search strategies (e.g. Evolutionary vs. Monte Carlo search). Below, we provide a non-exhaustive overview of recent advances in APO to give readers a taste of this rapidly evolving field.

Ma et al.~\cite{10.5555/3666122.3667992} uses a strategy based on the greedy search. Gradient-based approaches such as ProTeGI~\cite{pryzant2023automatic}, MAPO~\cite{li2024mapo}, TextGrad~\cite{yuksekgonul2025optimizing}, etc. use gradient descent-like algorithms to optimize prompt embeddings according to a predefined performance objective~\cite{wen2023hard}. Starting from a human prompt, these methods learn continuous valued embeddings or prompt proposals using the gradient directions. On the other hand, approaches such as InstructZero~\cite{chen2024instructzero} optimize low-dimensional ``soft prompts" using LLMs instead of directly optimizing discrete instructions. They balance exploitation vs exploration in this space using Bayesian Optimization techniques. In terms of discrete optimization, approaches such as EvoPrompt~\cite{zhang2023evoprompt} circumvent access to gradients and internal parameters of LLMs using Evolutionary Algorithms to generate candidate prompts, thus making them more general. Another approach to the same problem leverages in-context learning. For example, Automated Prompt Engineer (APE)~\cite{zhou2022ape} generates prompts based on a few real-life examples, tailoring them to the task without needing extensive training data. APE treats the instruction as a program and optimizes it by searching through a pool of instruction candidates proposed by an LLM to maximize a chosen score function. A complementary approach to APO uses reinforcement learning-based strategies such as OIRL~\cite{sun2023query}, which model the interaction between the query-prompt pair via a reward model for proposing and evaluating candidate prompts suited for arithmetic reasoning tasks. In contrast, meta-Prompting~\cite{suzgun2024meta} uses structural and syntactical aspects of tasks to create general prompts that guide the generation of task-specific prompts. In lieu of template-based approaches, DSPy~\cite{khattab2023dspy} frames prompt engineering as a declarative, compiler-driven optimization task. DSPy poses prompting as a computational graph pipeline for text transformation, where LLMs are invoked through parameterized self-improving modules. This provides great flexibility across diverse tasks given a computational budget.

APO techniques have been demonstrated across a wide variety of NLP tasks~\cite{zhou2022ape} such as text classification, question answering, zero-shot learning, text generation, summarization, sentiment analysis, paraphrasing, code generation, and interactive dialogue systems. Going one step further, knowledge graph construction is an essential but complex NLP task with great relevance to data lakehouses because it enhances the usability, accessibility, and interoperability of the data stored within them. While KG-construction could greatly benefit from APO techniques, its application for triple extraction from text has not been studied before to the best of our knowledge.

\section{Experiment Benchmarking Setup}
\label{sec:exp}
\subsection{Task}
Let $\mathbf{x} = (x_{1}, x_{2}, \dots, x_{n})$ denote a passage of natural language text with a list of terms, and let  $\mathcal{R} = \{r_{1}, r_{2}, \dots, r_{m}\}$ be a predefined set of canonical relation types.  The goal of the triple extraction task is to identify: (a) A set of entity mentions $\mathcal{E} = \{e_{1}, e_{2}, \dots, e_{k}\}$ from the text, (b) a subset of relations $r \in \mathcal{R}$ expressed in the text, and (c) a set of directed triples $\bigl\{(e_{i},\,r,\,e_{j}) \mid e_{i}, e_{j} \in \mathcal{E},\; r \in \mathcal{R}\bigr\}$ that are represents the semantic relationships between the entity pairs that are mentioned in text. An example of the task, along with inputs and outputs, is illustrated in Figure~\ref{fig:task_exp}.

\subsection{Datasets}

We used SynthIE and REBEL datasets in our experiments.

\medskip
\noindent \textbf{SynthIE}~\cite{josifoski-etal-2023-exploiting} is a relation extraction benchmark generated using subgraphs of Wikidata and a synthetic data generation strategy. We used the \textit{synthie\_text} (generated using \textit{text-davinci-003}) for training, validation and testing automatic prompt optimizers. The dataset contains 888 relation types.

\medskip
\noindent \textbf{REBEL}~\cite{huguet-cabot-navigli-2021-rebel-relation} is a relation extraction benchmark generated using Wikipedia abstracts and aligned them to Wikidata triples using wikilinks to identify relation candidates that are further validated using a Natural Language Inference (NLI) model. The dataset contained 1079 relation types.

\subsection{Automatic Prompt Optimizers}
\label{sub-sec:3-apo}
\begin{figure}
  \centering
  \includegraphics[width=0.7\linewidth]{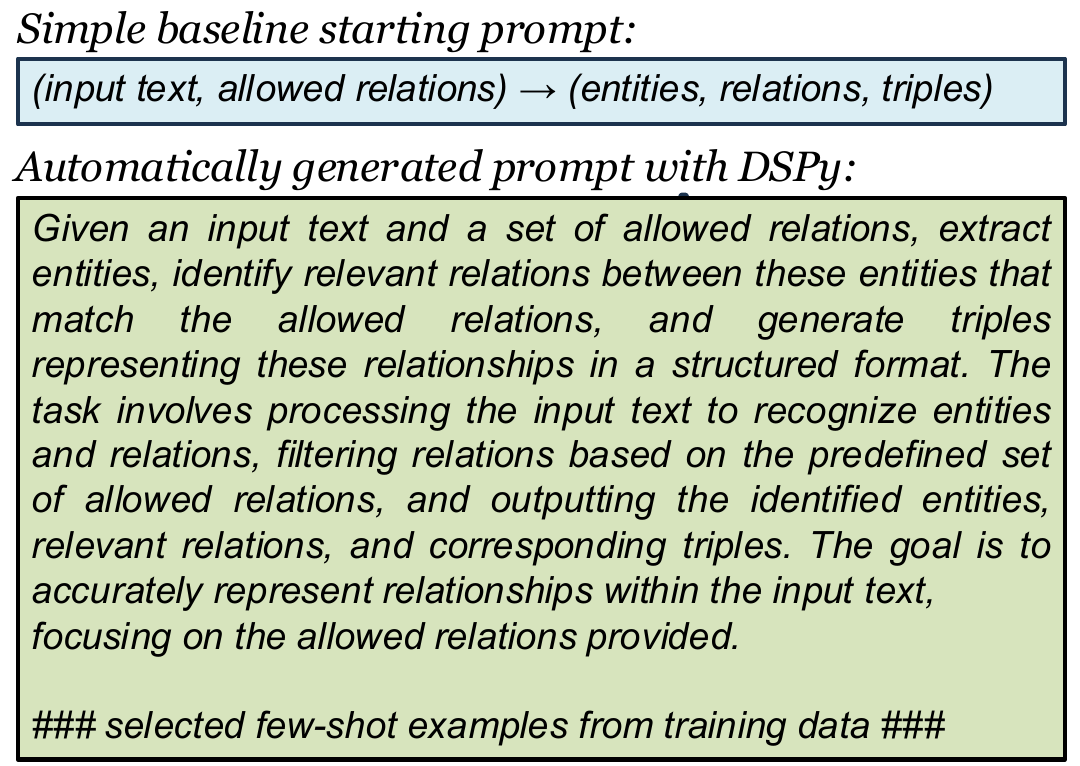}
  \caption{An example of an automatically optimized prompt using DSPy with the Llama 3.3-70B model and the SynthIE dataset as training/validation data.}
  \label{fig:ex_optimized_prompt}
\end{figure}

We used three approaches with distinct flavours of APO for triple extraction, detailed below:

\medskip
\noindent\textbf{DSPy}~\cite{khattab2023dspy} (Declarative Self-improving Python) is an open‑source toolkit for modular prompt design as programming with structured and declarative natural-language modules. DSPy allows specifying input/output behaviour as a signature and selecting a module (e.g., predict, Chain-Of-Thought, ReAct) to assign a strategy for invoking the LLM. DSPy natives parse the LLM output based on the pre-defined signature. It allows for the optimization of prompts using a training set, which is analyzed to create a `sub-dataset' and task descriptions, propose better prompts with LLM-generated instructions, and select effective few-shots examples. Specifically, we used MIPROv2 (Multiprompt Instruction PRoposal Optimizer Version 2)~\cite{opsahl-ong-etal-2024-optimizing} to optimize instructions and few-shot examples simultaneously as a joint search problem. In the rest of the paper, when DPSy is mentioned, it is reffering to DSPy with MIPROv2. It first bootstraps a set of few-shot example candidates, proposes instructions grounded in different task dynamics, and finds an optimized combination of these joint candidates using Bayesian optimization. Figure~\ref{fig:ex_optimized_prompt} shows an example of an automatically optimized prompt using DSPy.

\medskip
\noindent\textbf{APE}~\cite{zhou2022ape}
APE does not require any initial prompt. Given some examples of input and expected output, it generates several task instruction candidates, executes them using the target model, and selects the most appropriate instruction based on evaluation scores. By default, APE uses GPT-3 (text-davinci-002 to be more specific). We replace it with Granite-3.2-8B-Instruct, given its open-source availability. We used the forward generation template settings and randomly picked 10 examples from the SynthIE validation data to provide as few shots, and selected the generated prompt with the highest score based on the validation set of SynthIE.

\medskip
\noindent\textbf{TextGrad}~\cite{yuksekgonul2025optimizing}
Unlike statistical optimization in machine learning, which numerically estimates gradients for backpropagation, TextGrad leverages natural language feedback to iteratively improve LLM-generated prompts, effectively operating on the computational graph connecting the instruction to the output. We adopt the TextualGradientDescent optimizer from the official Pytorch implementation in ~\cite{yuksekgonul2025optimizing} with pre-set hyperparameters for triple extraction. The initial prompt is a basic instruction describing the task, required output format, allowed canonical relations, and an optional one-shot example input and output. The evaluation and training engines are chosen to be the LLama3-70b model. Training is performed for a maximum of 5 epochs, batch size of 16, allowing for early stopping using the validation set. The loss function driving the optimization is 1-F1, thus optimizing for both recall and precision simultaneously. Of the three approaches tried, text grad is the most computationally expensive and slow to train, as it requires at least two API calls for each element of the training set per epoch (one for the forward pass and a second for textual gradient estimation)

\subsection{Language Models}

The automatic prompt optimizers use language models to generate the prompts. In this work, we have used the following open-source large language models for prompt optimization and performing the triple extraction task:
\begin{itemize}
    \item Deepseek V3\footnote{\url{https://huggingface.co/deepseek-ai/DeepSeek-V3}}
    \item Qwen2.5-72B\footnote{\url{https://huggingface.co/Qwen/Qwen2.5-72B}}
    \item Meta Llama3.3-70B Instruct\footnote{\url{https://huggingface.co/meta-llama/Llama-3.3-70B-Instruct}}
    \item Mistral-8x22B\footnote{\url{https://huggingface.co/mistralai/Mixtral-8x22B-v0.1}}
    \item Microsoft Phi-4 14B\footnote{\url{https://huggingface.co/microsoft/phi-4}}
    \item IBM Granite3.2-8B Instruct\footnote{\url{https://huggingface.co/ibm-granite/granite-3.2-8b-instruct}}
\end{itemize}

\subsection{Evaluation Metrics}

\medskip
\noindent\textbf{Entity, Relation, Triple extraction (P, R, F1)} To evaluate entity, relation, and triple extraction, the extracted values are compared against the ground truth values to estimate the precision, recall, and F1 metrics. These metrics are calculated for each test-case, and the final reported values are the macro-averaged. 

\medskip
\noindent\textbf{Relation-Type Mean Accuracy} To analyze the results at the level of relation types, we compute the mean accuracy per relation type. It captures the average proportion of correctly identified relation instances for each specific relation type, computed against all instances of the given type. 

\section{Empirical Analysis and Insights}

In this section, we analyze the performance improvements obtained by automatic prompt optimization for triple extraction from text, and how certain aspects related to the task, such as the length of the text or the complexity of the schema and the system, i.e. LLM being used drive the performance.

In scenarios under investigation in the following sections, default settings are maintained, as follows.

\medskip
\noindent\textbf{Default settings}: Prompting strategy: Predict (E-R-T), LLM: Llama3.3-70B, Number of relations per test case: 100, Test dataset: SynthIE, Optimizer: DSPy, Optimization metric: Triple F1 score.

\subsection{Prompting strategy}

\textbf{RQ1}: How do different prompting strategies influence the performance improvements in triple extraction?

\textbf{Varied setting}: Prompting strategy.
 
We curated six propmting-strategy variants that uses distinct styles to obtain the expected output. This includes (a) Predict with In-Context Learning, (b) Chain-Of-Thought, and (c) piplelining subtask prompts for Extract, Critique, and Refine. The expected output may be one of (i) Triple (T) or (ii) Entity, Relations, and Triples (E-R-T). The six resulting variants are in Table~\ref{tab:apo_prompt_strategy}. Each setting has a baseline prompt (see Table~\ref{tab:apo_prompt_strategy}) and is run through the prompt optimization process which proposes a refined prompt. 

Table~\ref{tab:prompt_strategy_full_result} outlines the performance improvements in terms of macro Precision, Recall, and F1 obtained by prompt optimization. We observe that all prompting strategies result in a performance gain, as seen in Table~\ref{tab:prompt_strategy_full_result}. Of these, Chain-Of-Thought asked to solely extract triples obtains the best triple extraction F1 of $0.73$. The highest overall performance improvement is seen in the case of the Extract, Critique, and Refine pipeline run on entities, relations, and triples i.e., ECR (E-R-T) output, with a $+16\%$ improvement in triple extraction F1. 

\begin{figure}
  \centering
  \includegraphics[width=\linewidth]{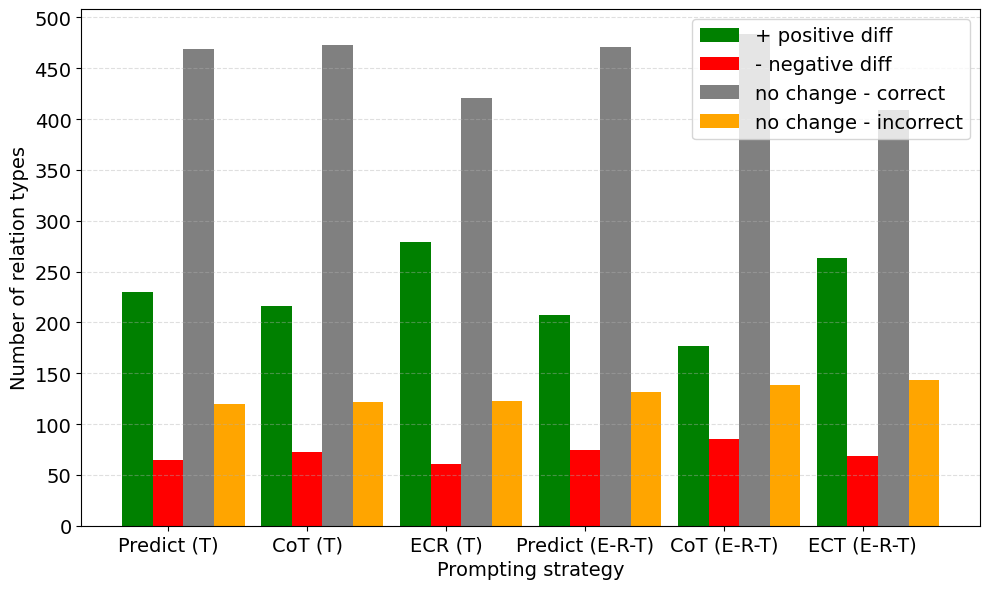}
  \caption{Number of relations types with +ve / -ve / 0 differences in accuracy after prompt optimization.}
  \label{fig:prompt_strategy_pref_diff_bar}
\end{figure}

Figure~\ref{fig:prompt_strategy_pref_diff_bar} shows the performance improvement at the relation-type level between the baseline prompt and the optimized prompt. It separates the number of relations that saw an increase in the mean accuracy (+ positive diff), a decrease in mean accuracy (- negative diff), and no change in mean accuracy before and after the prompt optimization process. The relation types that didn't have a change are further separated into the incorrect ones (i.e., mean accuracy 0) and the correct ones. It shows that in all prompting strategies, there are more relation types having a performance improvement (green) than a performance degradation (red) in Figure~\ref{fig:prompt_strategy_pref_diff_bar}. It also suggests that the optimized prompts retain the majority of correct relations (grey) from the baseline prompts.

\begin{figure}[htbp]
  \centering
  \begin{subfigure}{0.8\linewidth}
    \centering
    \includegraphics[width=\linewidth]{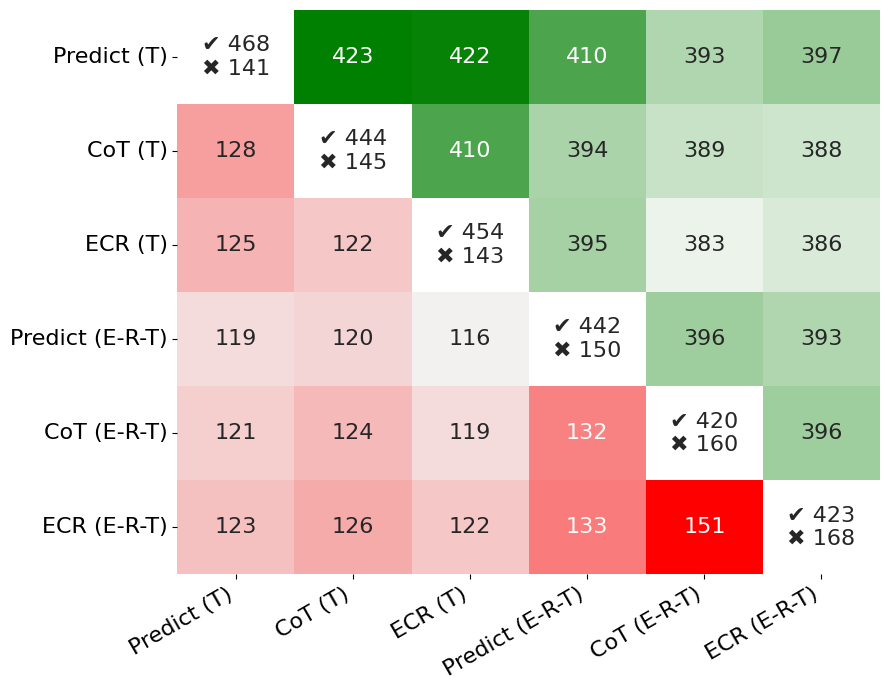}
    \caption{Optimized Prompts}
    \label{fig:ex1}
  \end{subfigure}
  \vspace{3em}
  \begin{subfigure}{0.8\linewidth}
    \centering
    \includegraphics[width=\linewidth]{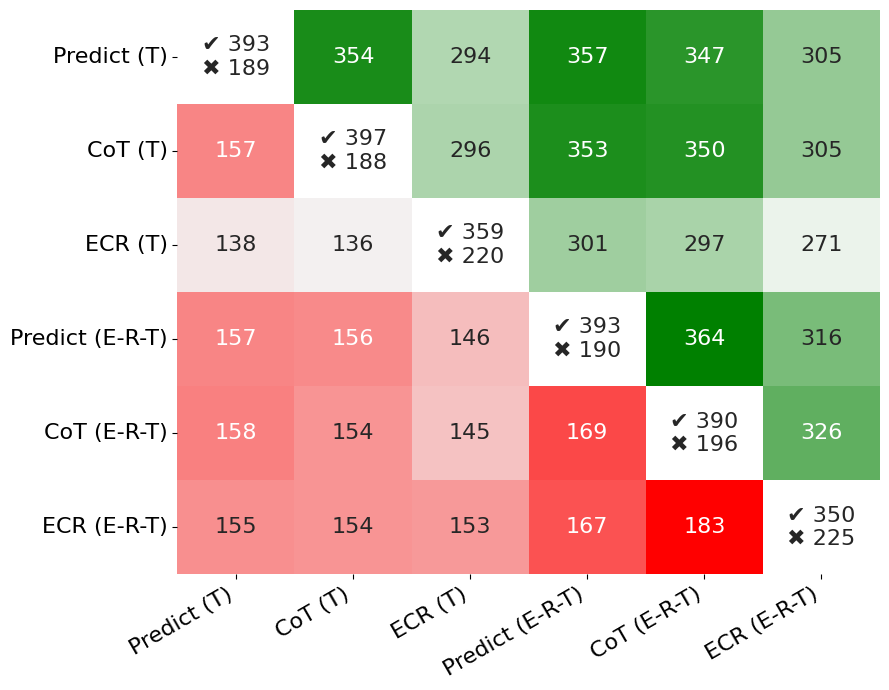}
    \caption{Baseline Prompts}
    \label{fig:ex2}
  \end{subfigure}
  \caption{Overlap of fully correct relation types, i.e., mean accuracy 1.0 (Upper, Green) and fully incorrect relation types, mean accuracy 0.0 (Lower, Red) of different baseline and optimized prompts. Diagonal shows the total count of fully correct (\ding{51}) and fully incorrect (\ding{55}) relation types for each optimized prompt.}
  \label{fig:prompt_strategy_rel_type_headmaps}
\end{figure}

Similarly, in Figure~\ref{fig:prompt_strategy_rel_type_headmaps}, we analyze if all approaches find the same types of relations easier to extract (extract them correctly always) and harder to extract (extract them incorrectly always). Each heatmap shows the overlap of fully correct relation types (Upper Right, Green) and fully incorrect relation types (Lower Left, Red) among the different prompting strategies for the baseline prompts and optimized prompts. The fully correct Relation-Types refer to the Relation-Types with a mean accuracy of 1.0, and fully incorrect Relation-Types refer to relations with a mean accuracy of 0. The diagonal shows the total number of fully correct and incorrect Relation-Types for each approach. We observe high overlap between different approaches; for example, between Predict (T) and Cot (T), there is a $95\%$ $(423/444)$ overlap in fully correct relations and a 88\% (128/145) overlap in fully incorrect relation types.

\begin{table}[ht]
\caption{Different prompting strategies for applying APO.}
\label{tab:apo_prompt_strategy}
\small
\setlength{\tabcolsep}{8pt}
\begin{tabular}{p{1cm} p{6cm} >{\raggedright\arraybackslash} p{8cm}}
\toprule
\textbf{APO} & \textbf{Description} & \textbf{Baseline Prompt} \\
\midrule
Predict (T) & Extracts triples from input text using a given list of allowed relations & \texttt{(input text, allowed relations) $\rightarrow$ (triples)} \\
\midrule
\makecell{CoT\\(T)} & Asks the model to think step-by-step and provide a reasoning in addition to the triples. & \texttt{(input text, allowed relations) $\rightarrow$ (reasoning, 
 triples)} \\
\midrule
\makecell{ECR \\(T)} & A 3-step pipeline: (E)xtract, (C)ritique, and (R)efine triples with three distinct prompts. Each step passes its output to the next stage. &
\texttt{P1: (input text, allowed relations) $\rightarrow$ (triples)} \newline
\texttt{P2: (input text, allowed relations, extracted triples) $\rightarrow$ (triple critique)} \newline
\texttt{P3: (input text, allowed relations, extracted triples, triple critique)  $\rightarrow$ (refined triples)} \\
\midrule
Predict (E-R-T) & Similar to Predict (T) but also explicitly extracts lists of entities and relations. & \texttt{(input text, allowed relations) $\rightarrow$ (entities, relations, triples)} \\
\midrule
\makecell{CoT\\ (E-R-T)} & Similar to CoT (T) but also explicitly extracts lists of entities and relations. & \texttt{(input text, allowed relations) $\rightarrow$ (reasoning, entities, relations, triples)} \\
\midrule
\makecell{ECR\\ (E-R-T)} & A 3-step pipeline 
 (E)xtract, (C)ritique, and (R)efine similar to ECR (T) but also explicitly extracts lists of entities and relations in addition to the triples. All outputs of the previous step is passed as input to the next step. & \texttt{P1: (input text, allowed relations) $\rightarrow$ (entities, relations, triples)} \newline \texttt{P2: ... $\rightarrow$ (entity critique, relation critique, triple critique)} \newline \texttt{P3: ... $\rightarrow$ (refined entities, refined relations, refined triples)}\\
\bottomrule
\end{tabular}
\end{table}

\begin{table}[ht]
\caption{Precision (P), Recall (R) , F1  metrics for Entity, Relation, Triple extractions using different prompting strategies. These experiments use Llama-3.3 70B as the LLM with $100$ allowed relations per test case, evaluated on SynthIE small test set.  White rows represent the baseline, and grey rows represent the optimized prompts.}
\label{tab:prompt_strategy_full_result}
\small
\setlength{\tabcolsep}{14pt}
\begin{tabular}{lccc|ccc|ccc}
\toprule
\textbf{\makecell{Prompting\\Strategy}} & \multicolumn{3}{c|}{\textbf{Entity}} & \multicolumn{3}{c|}{\textbf{Relation}} & \multicolumn{3}{c}{\textbf{Triple}} \\
\cmidrule(lr){2-4} \cmidrule(lr){5-7} \cmidrule(lr){8-10}
& P & R & F1 & P & R & F1 & P & R & F1 \\
\midrule
\multirow{2}{*}{Predict (T)} 
  & - & - & - & - & - & - & 0.69 & 0.60 & 0.63 \\
  & - & - & - 
  & - & - & - 
  & \cellcolor{gray!15}0.73 & \cellcolor{gray!15}0.73 & \cellcolor{gray!15}0.72 \\ \midrule
\multirow{2}{*}{CoT (T)} 
  & - & - & - & - & - & - & 0.71 & 0.60 & 0.64 \\
  & - & - & - 
  & - & - & - 
  & \cellcolor{gray!15}0.76 & \cellcolor{gray!15}0.72 & \cellcolor{gray!15}0.73 \\ \midrule
\multirow{2}{*}{ECR (T)} 
  & - & - & - & - & - & - & 0.60 & 0.59 & 0.59 \\
  & - & - & - 
  & - & - & - 
  & \cellcolor{gray!15}0.74 & \cellcolor{gray!15}0.71 & \cellcolor{gray!15}0.72 \\ \midrule
\multirow{2}{*}{\makecell{Predict\\(E-R-T)}} 
  & 0.98 & 0.90 & 0.93 & 0.80 & 0.66 & 0.71 & 0.68 & 0.58 & 0.62 \\
  & \cellcolor{gray!15}0.98 & \cellcolor{gray!15}0.94 & \cellcolor{gray!15}0.96 
  & \cellcolor{gray!15}0.80 & \cellcolor{gray!15}0.77 & \cellcolor{gray!15}0.77 
  & \cellcolor{gray!15}0.74 & \cellcolor{gray!15}0.71 & \cellcolor{gray!15}0.72 \\ \midrule
\multirow{2}{*}{\makecell{CoT\\ (E-R-T)}} 
  & 0.98 & 0.91 & 0.94 & 0.81 & 0.68 & 0.72 & 0.69 & 0.59 & 0.63 \\
  & \cellcolor{gray!15}0.98 & \cellcolor{gray!15}0.94 & \cellcolor{gray!15}0.95 
  & \cellcolor{gray!15}0.81 & \cellcolor{gray!15}0.78 & \cellcolor{gray!15}0.78 
  & \cellcolor{gray!15}0.74 & \cellcolor{gray!15}0.70 & \cellcolor{gray!15}0.71 \\ \midrule
\multirow{2}{*}{\makecell{ECR\\(E-R-T)}} 
  & 0.95 & 0.94 & 0.94 & 0.62 & 0.65 & 0.62 & 0.52 & 0.56 & 0.53 \\
  & \cellcolor{gray!15} 0.98 & \cellcolor{gray!15} 0.94 & \cellcolor{gray!15} 0.96 
  & \cellcolor{gray!15} 0.80 & \cellcolor{gray!15} 0.76 & \cellcolor{gray!15}0.77 
  & \cellcolor{gray!15} 0.72 & \cellcolor{gray!15}0.68 & \cellcolor{gray!15}0.69 \\
\bottomrule
\end{tabular}
\vspace{2em}
\end{table}

\subsection{Prompt Generation LLM and Triple Extraction LLM}

\textbf{RQ2}: How do different LLMs influence the performance improvements of automatic prompt optimisation techniques?

\noindent\textbf{Varied setting}: LLM used for prompt generation and testing.

We performed a number of experiments using DSPy to probe the research question above and report results in Table~\ref{tab:llm_results}. The length of the best prompts generated using Deepseek V3, Qwen2.5-72B, Llama3.3-70B, Mistral-8x22B, Phi-4 14B and Granite3.2-8B are $68$, 81, $77$, $58$, $268$ and $125$ words long, respectively.

Note that the results achieved by all models are similar in performance for entity extraction, regardless of the exact model used to generate prompts. However, larger models (Deepseek V3, Qwen2.5-72B, and Llama3.3-70B) result in significantly better results for relation (and triple) extraction.  Secondly, the best prompt generated using Llama3.3-70B is robust enough to obtain comparable results to those of the best prompt generated using other models, when testing with those. For example, with prompt generated using Llama3.3-70B and Phi-4 14B and tested using Phi-4 14B, we obtained $0.62$ an $0.61$ F1-scores respectively for triple extraction. Similarly, prompts generated with Granite3.2-8B and Llama3.3-70B obtain the same F1-scores of $0.54$. It might imply that smaller models are capable of generating equally effective prompts compared to larger models. Furthermore, we observe that the choice of LLM during inference has a significant impact on the performance, as we obtain better results using larger models. This may imply that the choice of LLM at inference is more important than the choice of LLM that generates prompts.

\begin{table}[ht]
\caption{Results for using different LLMs for automatic prompt generation and triple extraction tasks. Each model contains results with a baseline (no prompt optimization), a prompt generated using the same model, and a prompt generated from a different model. Models are ordered by parameter size.}
\label{tab:llm_results}
\small
\begin{tabular}{l | l | p{0.8cm} p{0.8cm} p{0.8cm} | p{0.8cm} p{0.8cm} p{0.8cm} | p{0.8cm} p{0.8cm} p{0.8cm}}
\toprule
\multirow{2}{*}{\textbf{\makecell{Validation\\ /Test LLM}}} & \multirow{2}{*}{\textbf{\makecell{Prompt \\ Gen. LLM}}} & \multicolumn{3}{c|}{\textbf{Entity}} & \multicolumn{3}{c|}{\textbf{Relation}} & \multicolumn{3}{c}{\textbf{Triple}} \\
\cline{3-11}
 & & \textbf{P} & \textbf{R} & \textbf{F1} & \textbf{P} & \textbf{R} & \textbf{F1} & \textbf{P} & \textbf{R} & \textbf{F1} \\
\midrule

\multirow{3}{*}{\makecell{Deepseek V3 \\671B}} & Baseline & 0.96 & 0.96 & 0.95 & 0.82 & 0.72 & 0.75 & 0.68 & 0.60 & 0.63 \\
 & Deepseek V3 & \cellcolor{gray!15} 0.98 & \cellcolor{gray!15} 0.92 & \cellcolor{gray!15} 0.94 & \cellcolor{gray!15} 0.83 & \cellcolor{gray!15} 0.78 & \cellcolor{gray!15} 0.80 & \cellcolor{gray!15} 0.76 & \cellcolor{gray!15} 0.71 & \cellcolor{gray!15} 0.73 \\
 & Llama3.3-70B & \cellcolor{gray!15}0.98 & \cellcolor{gray!15}0.95 & \cellcolor{gray!15}0.96 & \cellcolor{gray!15}0.84 & \cellcolor{gray!15}0.81 & \cellcolor{gray!15}0.82 & \cellcolor{gray!15}0.78 & \cellcolor{gray!15}0.75 & \cellcolor{gray!15}0.76 \\
\midrule

\multirow{3}{*}{Qwen2.5-72B} & Baseline & 0.97 & 0.95 & 0.96 & 0.80 & 0.72 & 0.75 & 0.71 & 0.64 & 0.66 \\
 & Qwen2.5-72B & \cellcolor{gray!15}0.98 & \cellcolor{gray!15}0.94 & \cellcolor{gray!15}0.96 & \cellcolor{gray!15}0.79 & \cellcolor{gray!15}0.77 & \cellcolor{gray!15}0.77 & \cellcolor{gray!15}0.73 & \cellcolor{gray!15}0.70 & \cellcolor{gray!15}0.71 \\
 & Llama3.3-70B & \cellcolor{gray!15}0.98 & \cellcolor{gray!15}0.95 & \cellcolor{gray!15}0.96 & \cellcolor{gray!15}0.81 & \cellcolor{gray!15}0.80 & \cellcolor{gray!15}0.80 & \cellcolor{gray!15}0.72 & \cellcolor{gray!15}0.72 & \cellcolor{gray!15}0.71 \\
\midrule

\multirow{3}{*}{Llama3.3-70B} & Baseline & 0.98 & 0.90 & 0.93 & 0.77 & 0.68 & 0.71 & 0.68 & 0.59 & 0.62 \\
 & Llama3.3-70B & \cellcolor{gray!15}0.98 & \cellcolor{gray!15}0.94 & \cellcolor{gray!15}0.96 & \cellcolor{gray!15}0.80 & \cellcolor{gray!15}0.77 & \cellcolor{gray!15}0.77 & \cellcolor{gray!15}0.74 & \cellcolor{gray!15}0.71 & \cellcolor{gray!15}0.72 \\
 & Phi-4 14B & \cellcolor{gray!15}0.98 & \cellcolor{gray!15}0.92 & \cellcolor{gray!15}0.94 & \cellcolor{gray!15}0.83 & \cellcolor{gray!15}0.72 & \cellcolor{gray!15}0.76 & \cellcolor{gray!15}0.74 & \cellcolor{gray!15}0.65 & \cellcolor{gray!15}0.68 \\
\midrule

\multirow{3}{*}{Mistral-8x22B} & Baseline & 0.96 & 0.96 & 0.95 & 0.64 & 0.61 & 0.61 & 0.55 & 0.52 & 0.53 \\
 & Mistral-8x22B & \cellcolor{gray!15}0.98 & \cellcolor{gray!15}0.93 & \cellcolor{gray!15}0.95 & \cellcolor{gray!15}0.67 & \cellcolor{gray!15}0.64 & \cellcolor{gray!15}0.64 & \cellcolor{gray!15}0.61 & \cellcolor{gray!15}0.58 & \cellcolor{gray!15}0.58 \\
 & Llama3.3-70B & \cellcolor{gray!15}0.98 & \cellcolor{gray!15}0.94 & \cellcolor{gray!15}0.95 & \cellcolor{gray!15}0.69 & \cellcolor{gray!15}0.68 & \cellcolor{gray!15}0.67 & \cellcolor{gray!15}0.59 & \cellcolor{gray!15}0.58 & \cellcolor{gray!15}0.58 \\
\midrule

\multirow{3}{*}{Phi-4 14B} & Baseline & 0.95 & 0.97 & 0.95 & 0.66 & 0.55 & 0.59 & 0.54 & 0.48 & 0.50 \\
 & Phi-4 14B & \cellcolor{gray!15}0.98 & \cellcolor{gray!15}0.92 & \cellcolor{gray!15}0.95 & \cellcolor{gray!15}0.73 & \cellcolor{gray!15}0.67 & \cellcolor{gray!15}0.69 & \cellcolor{gray!15}0.65 & \cellcolor{gray!15}0.59 & \cellcolor{gray!15}0.61 \\
 & Llama3.3-70B & \cellcolor{gray!15}0.98 & \cellcolor{gray!15}0.92 & \cellcolor{gray!15}0.94 & \cellcolor{gray!15}0.74 & \cellcolor{gray!15}0.67 & \cellcolor{gray!15}0.69 & \cellcolor{gray!15}0.67 & \cellcolor{gray!15}0.60 & \cellcolor{gray!15}0.62 \\
\midrule

\multirow{3}{*}{Granite3.2-8B} & Baseline & 0.99 & 0.83 & 0.89 & 0.55 & 0.44 & 0.47 & 0.47 & 0.37 & 0.40 \\
 & Granite3.2-8B & \cellcolor{gray!15}0.98 & \cellcolor{gray!15}0.90 & \cellcolor{gray!15}0.93 & \cellcolor{gray!15}0.66 & \cellcolor{gray!15}0.60 & \cellcolor{gray!15}0.62 & \cellcolor{gray!15}0.58 & \cellcolor{gray!15}0.52 & \cellcolor{gray!15}0.54 \\
 % & Granite3.2-8B (APE) & \cellcolor{gray!15}0.98 & \cellcolor{gray!15}0.91 & \cellcolor{gray!15}0.93 & \cellcolor{gray!15}0.58 & \cellcolor{gray!15}0.58 & \cellcolor{gray!15}0.57 & \cellcolor{gray!15}0.51 & \cellcolor{gray!15}0.49 & \cellcolor{gray!15}0.49 \\
 & Llama3.3-70B & \cellcolor{gray!15}0.98 & \cellcolor{gray!15}0.90 & \cellcolor{gray!15}0.93 & \cellcolor{gray!15}0.67 & \cellcolor{gray!15}0.59 & \cellcolor{gray!15}0.61 & \cellcolor{gray!15}0.59 & \cellcolor{gray!15}0.52 & \cellcolor{gray!15}0.54 \\

\bottomrule
\end{tabular}
\vspace{2em}
\end{table}

\subsection{Schema Complexity}

\textbf{RQ3}: What is the impact of schema complexity (i.e., number of allowed relations) on relation extraction performance? Are optimised prompts more robust to changes in schema complexity?

\noindent\textbf{Varied setting}: Number of relations per test case.

To analyze this research question, we change the number of allowed relations that we include in the prompt. When a larger number of relations are included in the input text, we expect that the increased search space makes the task slightly harder for the systems with the increase in relation types. We examine eight settings with the number of allowed relation types ranging from $100-800$. 

\begin{figure}
  \centering
  \includegraphics[width=\linewidth]{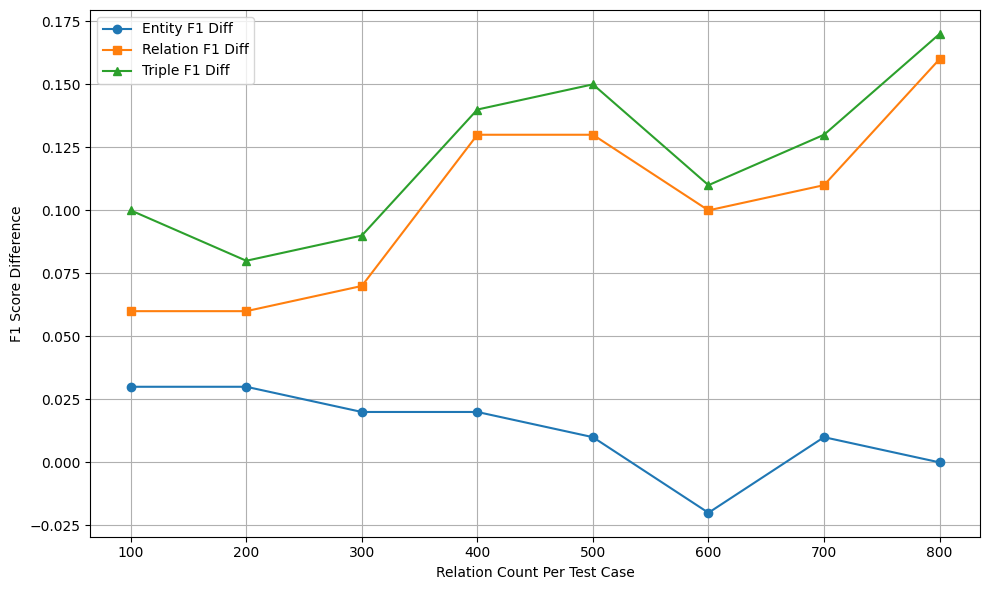}
  \caption{Difference in Macro F1 Scores (Optimized - Baseline) at different number of allowed relations. F1 is taken as the macro average of all test cases.}
  \label{fig:rel_count_res_by_test_case}
\end{figure}

Table~\ref{tab:rel_count_eval} illustrates the impact on Precision, Recall, and F1 metrics for Entity, Relation, and Triple extraction. As seen, the performance of relation and triple extraction drops significantly with an increased number of allowed relations, suggesting an increase in task complexity. Figure~\ref{fig:rel_count_res_by_test_case} displays the performance difference between the baseline and optimized prompt at the level of test-case while varying allowed relations 100 to 800. Interestingly, the F1 score for entity extraction reduces with the optimized prompt. For instance, in the setting where there are $600$ relations, it drops from $93\%$ to $91\%$. Hence, entity extraction scores do not show the same trend as in relation and triple extraction.

\begin{table}[ht]
\caption{Precision (P), Recall(R), F1 metrics for entity, relation, triple extractions across varying numbers of relation types per each test case. These experiments use LLAMA 3.3 70B as the LLM with Predict (E-R-T) as the prompting strategy, evaluated on the SynthIE small test set.White rows represent the baseline, and grey rows represent the optimised prompts.}
\label{tab:rel_count_eval}
\small
\setlength{\tabcolsep}{14pt}
\begin{tabular}{c|ccc|ccc|ccc}
\toprule
\textbf{\makecell{Relation\\Count\\Per \\Test Case}} & \multicolumn{3}{c|}{\textbf{Entity}} & \multicolumn{3}{c|}{\textbf{Relation}} & \multicolumn{3}{c}{\textbf{Triple}} \\
\cmidrule(lr){2-4} \cmidrule(lr){5-7} \cmidrule(lr){8-10}
 & P & R & F1 & P & R & F1 & P & R & F1 \\
\midrule
rel 100 & 0.98 & 0.90 & 0.93 & 0.77 & 0.68 & 0.71 & 0.68 & 0.58 & 0.62 \\
     & \cellcolor{gray!15}0.98 & \cellcolor{gray!15}0.94 & \cellcolor{gray!15}0.96 
     & \cellcolor{gray!15}0.80 & \cellcolor{gray!15}0.77 & \cellcolor{gray!15}0.77 
     & \cellcolor{gray!15}0.74 & \cellcolor{gray!15}0.71 & \cellcolor{gray!15}0.72 \\
\midrule
rel 200 & 0.98 & 0.90 & 0.93 & 0.75 & 0.65 & 0.68 & 0.66 & 0.57 & 0.60 \\
     & \cellcolor{gray!15}0.98 & \cellcolor{gray!15}0.94 & \cellcolor{gray!15}0.96 
     & \cellcolor{gray!15}0.77 & \cellcolor{gray!15}0.73 & \cellcolor{gray!15}0.74 
     & \cellcolor{gray!15}0.71 & \cellcolor{gray!15}0.67 & \cellcolor{gray!15}0.68 \\ \midrule
rel 300 & 0.98 & 0.90 & 0.93 & 0.74 & 0.63 & 0.67 & 0.65 & 0.55 & 0.59 \\
     & \cellcolor{gray!15}0.98 & \cellcolor{gray!15}0.94 & \cellcolor{gray!15}0.95 
     & \cellcolor{gray!15}0.77 & \cellcolor{gray!15}0.72 & \cellcolor{gray!15}0.74 
     & \cellcolor{gray!15}0.71 & \cellcolor{gray!15}0.66 & \cellcolor{gray!15}0.68 \\ \midrule
rel 400 & 0.98 & 0.90 & 0.93 & 0.62 & 0.48 & 0.53 & 0.54 & 0.42 & 0.46 \\
     & \cellcolor{gray!15}0.97 & \cellcolor{gray!15}0.93 & \cellcolor{gray!15}0.95 
     & \cellcolor{gray!15}0.69 & \cellcolor{gray!15}0.64 & \cellcolor{gray!15}0.66 
     & \cellcolor{gray!15}0.63 & \cellcolor{gray!15}0.59 & \cellcolor{gray!15}0.60 \\ \midrule
rel 500 & 0.98 & 0.90 & 0.93 & 0.62 & 0.47 & 0.52 & 0.54 & 0.41 & 0.45 \\
     & \cellcolor{gray!15}0.97 & \cellcolor{gray!15}0.93 & \cellcolor{gray!15}0.94 
     & \cellcolor{gray!15}0.69 & \cellcolor{gray!15}0.64 & \cellcolor{gray!15}0.65 
     & \cellcolor{gray!15}0.63 & \cellcolor{gray!15}0.58 & \cellcolor{gray!15}0.60 \\ \midrule
rel 600 & 0.98 & 0.90 & 0.93 & 0.66 & 0.51 & 0.56 & 0.58 & 0.44 & 0.49 \\
     & \cellcolor{gray!15}0.92 & \cellcolor{gray!15}0.92 & \cellcolor{gray!15}0.91 
     & \cellcolor{gray!15}0.69 & \cellcolor{gray!15}0.66 & \cellcolor{gray!15}0.66 
     & \cellcolor{gray!15}0.62 & \cellcolor{gray!15}0.58 & \cellcolor{gray!15}0.60 \\ \midrule
rel 700 & 0.98 & 0.90 & 0.93 & 0.62 & 0.47 & 0.52 & 0.54 & 0.41 & 0.45 \\
     & \cellcolor{gray!15}0.96 & \cellcolor{gray!15}0.92 & \cellcolor{gray!15}0.94 
     & \cellcolor{gray!15}0.67 & \cellcolor{gray!15}0.62 & \cellcolor{gray!15}0.63 
     & \cellcolor{gray!15}0.61 & \cellcolor{gray!15}0.56 & \cellcolor{gray!15}0.58 \\ \midrule
rel 800 & 0.98 & 0.89 & 0.93 & 0.57 & 0.42 & 0.47 & 0.50 & 0.37 & 0.41 \\
     & \cellcolor{gray!15}0.95 & \cellcolor{gray!15}0.92 & \cellcolor{gray!15}0.93 
     & \cellcolor{gray!15}0.67 & \cellcolor{gray!15}0.62 & \cellcolor{gray!15}0.63 
     & \cellcolor{gray!15}0.61 & \cellcolor{gray!15}0.57 & \cellcolor{gray!15}0.58 \\ 
\bottomrule
\end{tabular}
\vspace{2em}
\end{table}

\begin{figure}
  \centering
  \includegraphics[width=\linewidth]{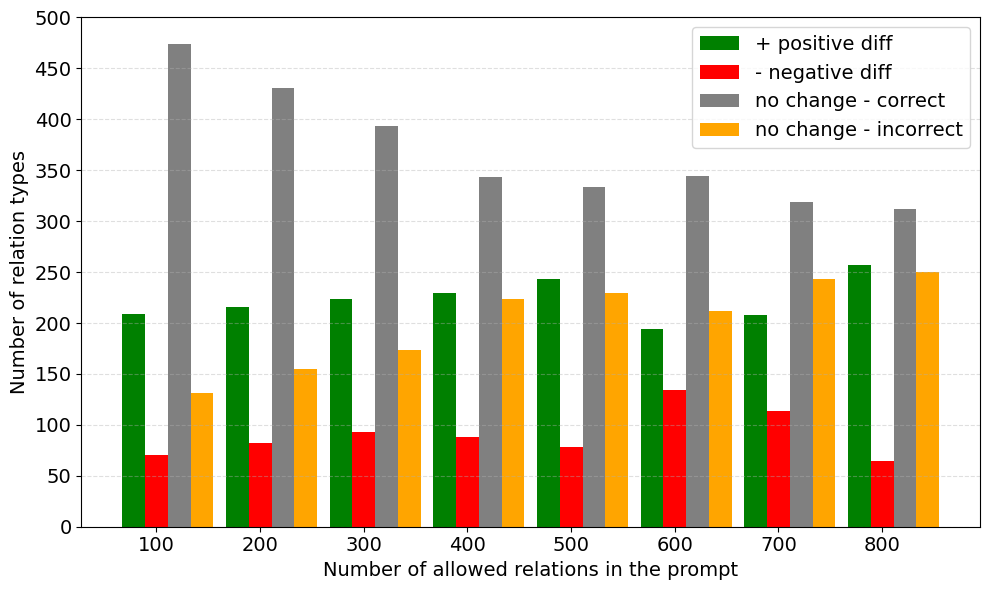}
  \caption{Number of relation types with +ve / -ve / 0 differences in mean accuracy after prompt optimization at each relation count (100 to 800) setting.}
  \label{fig:rel_count_res_by_rel_type}
\end{figure}
\setlength{\textfloatsep}{1pt} 

\subsection{Examining the effect of increased context length}

\textbf{RQ4:} How does APO perform at entity/relation/triple extraction as the size of the input text is increased?

\textbf{Varied setting}: Text length (1x, 5x, 10x test cases at a time).

In this experiment, we analyze the effect of increasing the context size of the input text on the performance of Automatic Prompt Optimization at entity/relation/triple extraction. Keeping the LLM fixed as Llama 3.3-70b, we systematically increase the length of the text in each example (in our small test set of 1500 examples) by augmenting it with text from the larger test corpus (10000 - 1500 examples). In the first setting, text examples that are most similar to those in our small test set are selected and appended, referred to as Related Augmentation. The second setting, named Adversarial Augmentation, repeats the same, but augments the data with the most dissimilar candidates. In the third setting, passages are randomly selected for augmentation which we refer to as Random Augmentation. We examine increasing the size of the input text 5 and 10 fold and compare it with the results from Section~5.1. Text similarity is determined using a sentence transformer model. The extraction results are summarized in Table~\ref{augment}. We also indicate the distinct relation types in each experiment, a quantitative proxy for the difficulty of the extraction task. The adversarial Augmentation and Random Augmentation have noticeably higher numbers of distinct relation types as compared to the related text case.

As expected, the performance of the baseline and trained DSPy modules degrades as the context length increases. The degradation is most severe in the cases where the text is augmented adversarially, with relation and triple recall suffering the most. Across the board, we observe that training for APO improves performance. We postulate that exposure to larger-sized text and especially their selection as few-shot examples, steers the optimization in a direction that helps better anticipate longer context. Improvements in performance are the largest in the case of Related Augmentation and the smallest in the Adversarial selection case. Finally, we notice that the performance in the Adversarial Augmentation case is noticeably worse than in the case of Random Augmentation, although the number of distinct relations is comparable. This suggests that the coherence of the text passage is an important factor influencing the difficulty of the extraction task.

\begin{table}[ht]
\caption{Quantifying the effect of increased context length on APO approaches for entity, relation and triple extraction. Three different settings are examined, where we merge (a) related text, (b) adversarially selected text, and (c) randomly selected text to each test example. White rows represent the baselines, and grey rows represent the optimized prompts. Numbers in the brackets indicate number of distinct relation types}
\label{augment}
\small
\setlength{\tabcolsep}{14pt}
\begin{tabular}{c | c c c | c c c | c c c}
\toprule
\multirow{2}{*}{\textbf{\makecell{Text size\\(Rel.~Types)}}} &
  \multicolumn{3}{c|}{\textbf{Entities}} &
  \multicolumn{3}{c|}{\textbf{Relations}} &
  \multicolumn{3}{c}{\textbf{Triples}} \\
\cmidrule(lr){2-4} \cmidrule(lr){5-7} \cmidrule(lr){8-10}
& \textbf{P} & \textbf{R} & \textbf{F1}
& \textbf{P} & \textbf{R} & \textbf{F1}
& \textbf{P} & \textbf{R} & \textbf{F1} \\
\midrule

\multicolumn{10}{c}{\textbf{(a) Merging Related Text}} \\
\midrule
\multirow{2}{*}{1x (3.10)} 
  & 0.98 & 0.90 & 0.93 
  & 0.77 & 0.68 & 0.71 
  & 0.68 & 0.59 & 0.62 \\
  & \cellcolor{gray!15}0.98 & \cellcolor{gray!15}0.94 & \cellcolor{gray!15}0.96 
  & \cellcolor{gray!15}0.80 & \cellcolor{gray!15}0.77 & \cellcolor{gray!15}0.77 
  & \cellcolor{gray!15}0.74 & \cellcolor{gray!15}0.71 & \cellcolor{gray!15}0.72 \\
\midrule
\multirow{2}{*}{5x (10.25)} 
  & 0.98 & 0.87 & 0.92 
  & 0.80 & 0.56 & 0.65 
  & 0.70 & 0.50 & 0.58 \\
  & \cellcolor{gray!15}0.99 & \cellcolor{gray!15}0.92 & \cellcolor{gray!15}0.95 
  & \cellcolor{gray!15}0.77 & \cellcolor{gray!15}0.67 & \cellcolor{gray!15}0.71 
  & \cellcolor{gray!15}0.71 & \cellcolor{gray!15}0.63 & \cellcolor{gray!15}0.66 \\
\midrule
\multirow{2}{*}{10x (17.11)} 
  & 0.99 & 0.86 & 0.91 
  & 0.82 & 0.47 & 0.59 
  & 0.70 & 0.43 & 0.53 \\
  & \cellcolor{gray!15}0.99 & \cellcolor{gray!15}0.92 & \cellcolor{gray!15}0.95 
  & \cellcolor{gray!15}0.79 & \cellcolor{gray!15}0.58 & \cellcolor{gray!15}0.67 
  & \cellcolor{gray!15}0.71 & \cellcolor{gray!15}0.55 & \cellcolor{gray!15}0.62 \\

\midrule
\multicolumn{10}{c}{\textbf{(b) Merging Adversarially Selected Text (most unrelated)}} \\
\midrule
\multirow{2}{*}{1x (3.10)} 
  & 0.98 & 0.90 & 0.93 
  & 0.77 & 0.68 & 0.71 
  & 0.68 & 0.59 & 0.62 \\
  & \cellcolor{gray!15}0.98 & \cellcolor{gray!15}0.94 & \cellcolor{gray!15}0.96 
  & \cellcolor{gray!15}0.80 & \cellcolor{gray!15}0.77 & \cellcolor{gray!15}0.77 
  & \cellcolor{gray!15}0.74 & \cellcolor{gray!15}0.71 & \cellcolor{gray!15}0.72 \\
\midrule
\multirow{2}{*}{5x (13.57)} 
  & 0.96 & 0.86 & 0.89 
  & 0.80 & 0.56 & 0.65 
  & 0.69 & 0.49 & 0.57 \\
  & \cellcolor{gray!15}0.99 & \cellcolor{gray!15}0.85 & \cellcolor{gray!15}0.91 
  & \cellcolor{gray!15}0.83 & \cellcolor{gray!15}0.58 & \cellcolor{gray!15}0.67 
  & \cellcolor{gray!15}0.72 & \cellcolor{gray!15}0.51 & \cellcolor{gray!15}0.59 \\
\midrule
\multirow{2}{*}{10x (23.89)} 
  & 0.96 & 0.85 & 0.89 
  & 0.83 & 0.48 & 0.60 
  & 0.72 & 0.43 & 0.54 \\
  & \cellcolor{gray!15}0.98 & \cellcolor{gray!15}0.93 & \cellcolor{gray!15}0.95 
  & \cellcolor{gray!15}0.83 & \cellcolor{gray!15}0.66 & \cellcolor{gray!15}0.73 
  & \cellcolor{gray!15}0.75 & \cellcolor{gray!15}0.61 & \cellcolor{gray!15}0.67 \\

\midrule
\multicolumn{10}{c}{\textbf{(c) Merging Randomly Selected Text}} \\
\midrule
\multirow{2}{*}{1x (3.10)} 
  & 0.98 & 0.90 & 0.93 
  & 0.77 & 0.68 & 0.71 
  & 0.68 & 0.59 & 0.62 \\
  & \cellcolor{gray!15}0.98 & \cellcolor{gray!15}0.94 & \cellcolor{gray!15}0.96 
  & \cellcolor{gray!15}0.80 & \cellcolor{gray!15}0.77 & \cellcolor{gray!15}0.77 
  & \cellcolor{gray!15}0.74 & \cellcolor{gray!15}0.71 & \cellcolor{gray!15}0.72 \\
\midrule
\multirow{2}{*}{5x (14.35)} 
  & 0.99 & 0.85 & 0.91 
  & 0.83 & 0.58 & 0.67 
  & 0.72 & 0.51 & 0.59 \\
  & \cellcolor{gray!15}0.98 & \cellcolor{gray!15}0.92 & \cellcolor{gray!15}0.95 
  & \cellcolor{gray!15}0.82 & \cellcolor{gray!15}0.68 & \cellcolor{gray!15}0.74 
  & \cellcolor{gray!15}0.74 & \cellcolor{gray!15}0.62 & \cellcolor{gray!15}0.67 \\
\midrule
\multirow{2}{*}{10x (26.20)} 
  & 0.98 & 0.84 & 0.90 
  & 0.85 & 0.51 & 0.63 
  & 0.73 & 0.45 & 0.55 \\
  & \cellcolor{gray!15}0.99 & \cellcolor{gray!15}0.91 & \cellcolor{gray!15}0.95 
  & \cellcolor{gray!15}0.84 & \cellcolor{gray!15}0.63 & \cellcolor{gray!15}0.72 
  & \cellcolor{gray!15}0.75 & \cellcolor{gray!15}0.57 & \cellcolor{gray!15}0.64 \\

\bottomrule
\end{tabular}
\vspace{2em}
\end{table}

\subsection{Train, Validation and Test Datasets}

\textbf{RQ5}: What is the impact of using train/validation splits from a different dataset for selecting few-shots and validating the generated prompts compared to using the train/validation splits from the dataset as the test split? 

\textbf{Varied setting}: Train/validation dataset, prompt strategy. 

DSPy uses a training set to select the few-shot candidates and create a task/dataset summary that can be used as input to the LLM to derive prompt instruction candidates. In this experiment, we explore the use of different combinations of training/validation and testing datasets from distinct datasets and examine how the results change compared to using the same dataset for training and testing. We examine this behaviour using Predict (T) and Predict (E-R-T) prompting strategies. 

In DSPy, few-shot candidates are selected from a training set and task/dataset summary is created. This is subsequently used as input for deriving prompt instruction candidates from an LLM. In this experiment, we explore the impact of using training/validation splits for different datasets compared to the test split. Specifically, we investigate how the results differ when the same dataset is used for both training and testing versus when different datasets are employed. We evaluate this behaviour using the Predict (T) and Predict (E-R-T) prompting strategies.

As shown in Table~\ref{tab:dataset_change}, the results demonstrate that using the same dataset for both few-shot selection and prompt validation significantly outperforms the case when distinct datasets are used for training/validation and testing. For instance, when SynthIE is used as the test set, the Predict (E-R-T) prompt improves triple F1 by $+0.01$ when trained on the REBEL dataset, compared to a $+0.08$ improvement when the same dataset is used for both training and testing. Similarly, with Predict (T) and SynthIE as the test set, using REBEL for training/validation negatively impacts the optimized prompt, reducing triple F1 by $-0.04$ points. When REBEL is used as the test set, a similar trend is observed, with a larger margin of improvement when the same dataset is used for both training and testing compared to when different datasets are employed. Our hypothesis is that using few-shot examples and dataset summaries generated from a dataset that differs from the final test introduces some bias, leading to a prompt that is suboptimal for the test set.

\begin{table}[ht]
\caption{Results for training and testing across two different triple extraction datasets.}
\label{tab:dataset_change}
\small
\setlength{\tabcolsep}{16pt}
\begin{tabular}{l l c | c c | c c c}
\toprule
\multirow{2}{*}{\textbf{\makecell{Test\\Set}}} & \multirow{2}{*}{\textbf{\makecell{Prompt\\Strategy}}} & \multirow{2}{*}{\textbf{\makecell{Train/Val\\Set}}} 
& \multirow{2}{*}{\textbf{\makecell{Entity \\ F1}}} & \multirow{2}{*}{\textbf{\makecell{Rel. \\F1}}} 
& \multicolumn{3}{c}{\textbf{Triple}} \\
\cmidrule(lr){6-8}
&  & & & &  \textbf{P} & \textbf{R} & \textbf{F1}\\
\midrule

\multirow{6}{*}{SynthIE} 
  &  & N/A     & - & - & 0.69 & 0.60 & 0.63 \\
  & \makecell{Predict \\(T)}            & SynthIE & \cellcolor{gray!15}- & \cellcolor{gray!15}- & \cellcolor{gray!15}0.73 & \cellcolor{gray!15}0.73 & \cellcolor{gray!15}0.72 \\
  &             & REBEL   & \cellcolor{gray!15}- & \cellcolor{gray!15}- & \cellcolor{gray!15}0.73 & \cellcolor{gray!15}0.52 & \cellcolor{gray!15}0.59 \\
\cmidrule(lr){2-8}
  &  & N/A     & 0.93 & 0.71 & 0.68 & 0.58 & 0.62 \\
  & \makecell{ Predict \\ (E-R-T)}                & SynthIE & \cellcolor{gray!15}0.95 & \cellcolor{gray!15}0.77 & \cellcolor{gray!15}0.72 & \cellcolor{gray!15}0.70 & \cellcolor{gray!15}0.70 \\
  &                 & REBEL   & \cellcolor{gray!15}0.91 & \cellcolor{gray!15}0.71 & \cellcolor{gray!15}0.71 & \cellcolor{gray!15}0.58 & \cellcolor{gray!15}0.63 \\
\midrule

\multirow{6}{*}{REBEL} 
  &  & N/A     & - & - & 0.24 & 0.29 & 0.23 \\
  & \makecell{Predict \\(T)} & REBEL   & \cellcolor{gray!15}- & \cellcolor{gray!15}- & \cellcolor{gray!15}0.37 & \cellcolor{gray!15}0.35 & \cellcolor{gray!15}0.33 \\
  &             & SynthIE & \cellcolor{gray!15}- & \cellcolor{gray!15}- & \cellcolor{gray!15}0.22 & \cellcolor{gray!15}0.37 & \cellcolor{gray!15}0.25 \\
\cmidrule(lr){2-8}
  &  & N/A     & 0.73 & 0.36 & 0.26 & 0.28 & 0.24 \\
  & \makecell{ Predict \\ (E-R-T)}               & REBEL   & \cellcolor{gray!15}0.77 & \cellcolor{gray!15}0.48 & \cellcolor{gray!15}0.34 & \cellcolor{gray!15}0.42 & \cellcolor{gray!15}0.35 \\
  &                 & SynthIE & \cellcolor{gray!15}0.74 & \cellcolor{gray!15}0.39 & \cellcolor{gray!15}0.24 & \cellcolor{gray!15}0.35 & \cellcolor{gray!15}0.26 \\
\bottomrule
\end{tabular}
\vspace{2em}
\end{table}

\subsection{Prompt Optimization Approach}

\textbf{RQ6}: What prompt optimization approach works best for triple extraction?

\textbf{Varied setting}: Prompt optimization approach.

We experiment with three prompt optimization systems for triple extraction, namely, DSPy,
TextGrad, and APE. As mentioned in Section \ref{sub-sec:3-apo}, these methods differ in how they approach the problem. We include the best possible prompt (as generated by the corresponding system) for each case and then input it to the Llama3.3-70B model at test time. As seen in Table \ref{tab:prompt_optimizer_results}, the APE system is very competitive with the other methods despite being more simplistic in formulation and not requiring an initial human prompt as input. However, since APE needs to venture a guess about the task without any initial prompt and only a couple of input-output pairs, the optimal instruction is found to be generic (in a linguistic and descriptive sense) and not as task-specific compared with the two other approaches. For example, the APE system generated the following  -

\begin{quote}
     Create a knowledge graph representation for the provided information, where each input is transformed into a list of dictionaries. Each dictionary should contain ``subject", ``object", and ``predicate" as keys, with their respective values being the surface form of the subject/object and predicate phrase.
\end{quote}

While the DSPy generated -
\begin{quote}
    Given a text and a predefined set of allowed relations, extract the entities mentioned in the text and determine the relationships between them based on the allowed relations. Identify the subject, object, and relation for each extracted relationship and represent them as triples. Ensure that the extracted entities and relations are accurately matched to the allowed relations, and provide a comprehensive list of entities, relations, and triples that reflect the semantic relationships present in the input text.
\end{quote}

And textgrad resulted in -
\begin{quote}
    Extract concise and directly related knowledge graph triples from the text, focusing on explicit entity relationships, and only utilize the specified allowed relations to maximize precision and recall, ensuring that the subject and object are directly mentioned in the text and the relation is clearly implied.
\end{quote}

According to the results, DSPy does have a small edge over the other two in performance.

\begin{table}[ht]
\caption{Results for generating prompts with different automatic prompt optimizers.}
\label{tab:prompt_optimizer_results}
\small
\setlength{\tabcolsep}{10pt}
\begin{tabular}{l | c c c | c c c | c c c }
\toprule
\multirow{2}{*}{\textbf{\makecell{Automatic \\ Prompt Optimizer}}} & \multicolumn{3}{c|}{\textbf{Entity}} & \multicolumn{3}{c|}{\textbf{Relation}} & \multicolumn{3}{c}{\textbf{Triple}} \\
\cmidrule(lr){2-10}
& \textbf{P} & \textbf{R} & \textbf{F1} & \textbf{P} & \textbf{R} & \textbf{F1} & \textbf{P} & \textbf{R} & \textbf{F1} \\
\midrule
Baseline
  & 0.98 & 0.90 & 0.93 & 0.77 & 0.68 & 0.71 & 0.68 & 0.59 & 0.62 \\
DSPy
  & \cellcolor{gray!15} 0.98 & \cellcolor{gray!15} 0.94 & \cellcolor{gray!15} 0.96 & \cellcolor{gray!15}0.80 & \cellcolor{gray!15}0.77 & \cellcolor{gray!15}0.77 & \cellcolor{gray!15}0.74 & \cellcolor{gray!15}0.71 & \cellcolor{gray!15}0.72 \\
APE
  & \cellcolor{gray!15}0.98 & \cellcolor{gray!15}0.94 & \cellcolor{gray!15}0.95 & \cellcolor{gray!15}0.79 & \cellcolor{gray!15}0.77 & \cellcolor{gray!15}0.76 & \cellcolor{gray!15}0.72 & \cellcolor{gray!15}0.67 & \cellcolor{gray!15}0.69 \\
TextGrad 
  & \cellcolor{gray!15}0.98 & \cellcolor{gray!15}0.93 & \cellcolor{gray!15}0.95 & \cellcolor{gray!15}0.79 & \cellcolor{gray!15}0.75 & \cellcolor{gray!15}0.76 & \cellcolor{gray!15}0.71 & \cellcolor{gray!15}0.65 & \cellcolor{gray!15}0.67 \\
\bottomrule
\end{tabular}
\vspace{3em}
\end{table}

\subsection{Optimization metric}
\textbf{RQ7}: What is the impact of the optimization metric on the performance of the optimized prompt? 

\textbf{Varied setting}: Prompt optimization metric.

Recall that DSPy selects the few-shot candidates from the training data and generates a set of instructions using an LLM. The candidate search space, including both the instructions and few-shot examples, is then evaluated using Bayesian optimization. The validation set is used to assess the candidates and guide the search strategy. During the process, a predefined optimization metric drives the optimization process, i.e. to guide the candidate selection and evaluate against the validation set.

To explore the impact of the optimization metric, we vary the metric used during training to select the optimal prompts. We aim to determine whether this variation influences the prompt’s performance at test time as evaluated against the test set. Specifically, we seek to verify whether a prompt optimized for F1, Precision, or Recall performs better on the corresponding metric during testing.

Table~\ref{tab:optimization_metrics} indicates the results for different settings with the three distinct optimization metrics. From the results, we observe that when triple precision is used as a metric, the triple precision on the test set improves slightly ($0.75$ compared to $0.71$ and $0.72$), as anticipated. Similarly, Recall also drives improvement in test Recall ($0.71$ compared to $0.66$ and $0.70$). Overall, this performance profile suggests that the optimization metric indeed influences the prompt’s behavior, but rather moderately in the specific setting evaluated. In addition, we manually inspect the three generated prompts to identify any linguistic signals correlating with the metrics. However, no notable instructional differences were found that that drive the LLM in a particular direction.

\begin{table}[ht!]
\caption{Results for optimizing the prompt using different optimization metrics.}
\label{tab:optimization_metrics}
\small
\setlength{\tabcolsep}{12pt}
\begin{tabular}{l | c c c | c c c | c c c }
\toprule
\multirow{2}{*}{\textbf{\makecell{Optimization\\ Metric}}} & \multicolumn{3}{c|}{\textbf{Entity}} & \multicolumn{3}{c|}{\textbf{Relation}} & \multicolumn{3}{c}{\textbf{Triple}} \\
\cmidrule(lr){2-10}
& \textbf{P} & \textbf{R} & \textbf{F1} & \textbf{P} & \textbf{R} & \textbf{F1} & \textbf{P} & \textbf{R} & \textbf{F1} \\
\midrule
{Baseline} 
  & 0.98 & 0.90 & 0.93 & 0.80 & 0.66 & 0.71 & 0.68 & 0.58 & 0.62 \\
Triple F1
  & \cellcolor{gray!15}0.98 & \cellcolor{gray!15}0.94 & \cellcolor{gray!15}0.95 & \cellcolor{gray!15}0.79 & \cellcolor{gray!15}0.77 & \cellcolor{gray!15}0.77 & \cellcolor{gray!15}0.72 & \cellcolor{gray!15}0.70 & \cellcolor{gray!15}0.70 \\
Triple Precision
  & \cellcolor{gray!15}0.98 & \cellcolor{gray!15}0.92 & \cellcolor{gray!15}0.95 & \cellcolor{gray!15}0.83 & \cellcolor{gray!15}0.72 & \cellcolor{gray!15}0.76 & \cellcolor{gray!15}0.75 & \cellcolor{gray!15}0.66 & \cellcolor{gray!15}0.69 \\
Triple Recall
  & \cellcolor{gray!15}0.98 & \cellcolor{gray!15}0.95 & \cellcolor{gray!15}0.96 & \cellcolor{gray!15}0.78 & \cellcolor{gray!15}0.79 & \cellcolor{gray!15}0.77 & \cellcolor{gray!15}0.71 & \cellcolor{gray!15}0.71 & \cellcolor{gray!15}0.70 \\
\bottomrule
\end{tabular}
\end{table}

\subsection{Training Computation Cost}
\textbf{RQ8}: What is the impact of increasing the number of LLM calls during optimization on the final performance?

\textbf{Varied setting}: APO Hyperparameters such as number of candidates, trials, mini-batch validation size, validation set size.

In this study, we define the computational cost of training as the number of LLM calls made during the process. In DSPy, the number of LLM calls is determined by several hyperparameters, including: the number of candidate prompts generated, the number of optimization trials conducted, the size of the mini-batch evaluated in each trial, the \texttt{full\_eval\_steps} (i.e., the number of mini-batch trials after which a full evaluation of the validation set is performed), and the size of the validation set. To assess the impact of these factors on the final performance of the generated prompt, we conducted experiments where we varied these hyperparameters.

\begin{figure}
  \centering
  \includegraphics[width=\linewidth]{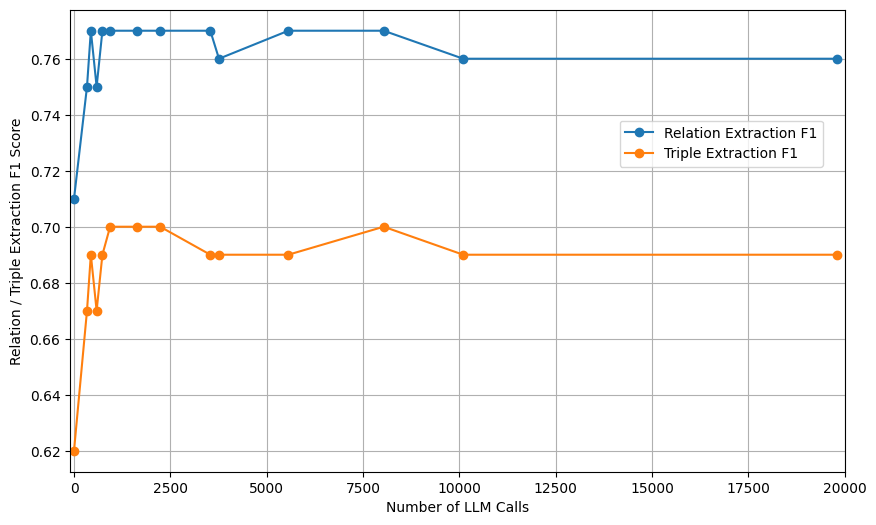}
  \caption{Training computation cost (in LLM calls) vs the relation and triple extraction F1.}
  \label{fig:computation_cost}
\vspace{2em}
\end{figure}

As shown in Figure~\ref{fig:computation_cost}, both relation extraction and triple extraction F1 scores stabilise after a relatively small number of LLM calls. The stability of these results can be numerically quantified via the Coefficient of Variation (CV) in the triple test metrics (i.e. the relative ratio of standard deviation to mean), with lower values implying that the variation is less pronounced (lower sensitivity to hyperparameter choices). Increasing the search space, by increasing the number of candidates and trials (${CV}_{\text{TripleR}} = 0.02, {CV}_{\text{TripleP}} =0.026, {CV}_{\text{TripleF1}} =0.019$), or the validation set size (${CV}_{\text{TripleR}} = 0.014, {CV}_{\text{TripleP}} =0.014, {CV}_{\text{TripleF1}} =0.036$), does not result in a drastic change in performance. These findings suggest that beyond a certain point, expanding the search space and increasing computational effort do not yield proportional gains in performance.

\section{Qualitative Analysis of Generated Prompts}

Three of the co-authors manually evaluated the top 22 prompts generated by Llama3.3-70B model using DSPy in different settings as described in Section \ref{sec:exp}. For each of these prompts, we asked ourselves \emph{``Would a human prompt engineer write this prompt and consider it an optimal or near-optimal for the triple extraction task, i.e., do you expect a good result when this prompt is provided to an LLM?"}. We used for 4 categories to mark the prompts -- \texttt{Good}, \texttt{Short}, \texttt{Ok} and \texttt{Verbose}. For these 22 prompt annotations, all three human annotators found that 3 prompts are \texttt{Verbose} and 8 prompts are \texttt{Good}. For the remaining 11, at least one (but not all 3) annotators marked 8 prompts as good.

The longest generated prompt marked "good" by all the annotators consists of 100 words -

\begin{quote}
    \small Given a piece of text and a set of allowed relations, identify the entities present in the text, determine the relationships between these entities based on the allowed relations, and construct triples in the form of subject-relation-object representing the extracted information. The goal is to accurately extract entities, relations, and triples from the input text, utilising the provided allowed relations to guide the relationship identification process. Ensure the output includes a list of identified entities, a list of relations found between these entities, and a list of triples that effectively capture the relationships between the entities in the text.
\end{quote}
Whereas the shortest prompt marked "good" by all the annotators consists of 77 words -

\begin{quote}
    \small Given a text and a predefined set of allowed relations, extract the entities mentioned in the text and determine the relationships between them based on the allowed relations. Identify the subject, object, and relation for each extracted relationship and represent them as triples. Ensure that the extracted entities and relations are accurately matched to the allowed relations, and provide a comprehensive list of entities, relations, and triples that reflect the semantic relationships present in the input text.
\end{quote}

Interesting, both the prompts yieleded similar results (when fed into the Llama3.3-70B) on the validation data - \emph{F1-Entities: $0.95$/F1-Relations: $0.77$/F1-Triples: $0.70$}.

\section{Conclusions}

In this empirical study, we explored the application of automatic prompt optimization for triple extraction via extensive experimental benchmarking. We evaluated different settings by changing (a) the prompting strategy, (b) the LLM being used for prompt optimization and task execution, (c) number of canonical relations in the schema (schema complexity), (d) the length and diversity of input text,  (e) the metric used to drive the prompt optimization, and (f) the dataset being used for training and testing. Based on the empirical results, automatic prompt optimization is a viable strategy for generating reasonable prompts for the triple extraction task, providing robust improvements against baseline prompts. The three prompt optimization strategies examine, namely, DSPy, APE, and TextGrad, consistently outperformed the baseline prompts in precision, recall, and F1 metrics for entity, relation, and triple extraction. Additionally, qualitative analysis of the generated prompts showed that a majority of the generated prompts were reasonably interpretable and in alignment with human intuition. Specifically, they provide explicit details of the task, desired output format, and some guidance on how to perform the task.

The experiments with different prompting styles showed that both simple inference prompts (predict) as well as advanced prompting with a Chain-of-Thought (Cot) rationale or a pipelined Extract-Critique-Refine (ECR) prompts provided improvements during prompt optimizations. These observations suggest that the automatic optimizers can adapt and propose effective refinements across the board in scenarios related to triple extraction. Analysis at the relation-type level showed that optimized prompts tend to retain or improve performance on the majority of relation-types present in the dataset, without losing performance on those relation-types that were already handled correctly by the baseline. This indicates that APO not only boosts overall macro-level metrics but also does so broadly across different relation types, suggesting that the improvements are generalizable, i.e. do not narrowly focus on a few types.

Experiments in increased schema-complexity (i.e., the number of allowed relation types) and input context length (1x, 5x, and 10x of the original input text) showed that automatic prompt optimization was considerably more beneficial the more challenging the scenario. When the schema became more complex with a larger number of relations (from 100 up to 800 relations), extraction difficulty increased, and both baseline and optimized prompt performance drops in terms of relation and triple F1. Nevertheless, the decline was relatively more significant for baseline prompts, implying that the optimized prompts were more resilient to increased schema-complexity. Increasingly long texts (by concatenating multiple test phrases together), both baseline and optimized prompt performance deteriorated, especially on relation and triple extraction (with relation extraction recall dropping due to challenges in identifying all relevant relations in a large blob of text). Optimized prompts can mitigate these context-length performance losses by still outperforming the baseline performance metrics for the single test case 1x all the way up to 10x. Descriptive optimized prompts with the right set of few-shot demonstrations seem to scale to larger input lengths quite gracefully.

\subsubsection*{\textbf{Limitations and Challenges}} When tested with different LLMs, it was observed that the triple extraction performance was mostly dominated by the capabilities of the task execution LLM rather than the LLM used to generate the prompts. The results imply that the choice of LLM at inference time seems more critical than the choice of LLM for prompt generation. In our experiments, prompts generated using a larger model (Llama 3.3-70B) were reasonably robust when applied to a different model (e.g. a 14B parameter model) on triple extraction. For example, the best prompt from Llama-70B tested on a smaller model achieved nearly the same triple F1 as the smaller model’s own optimized prompt. This suggests that there is some degree of ``model-agnostic prompt quality" maintained in the optimised prompts (typically a good prompt containing precise instructions that are generally useful).

When tested with multiple datasets, our results also highlight some limitations in cross-dataset prompt transferability. For example, when we used a different dataset’s training examples to optimize prompts (this simulates a shift in domain/data distribution), the improvements reduced significantly compared to using the examples from the same dataset. In particular, an optimized prompt trained on a secondary dataset yielded only marginal gains (on $\sim 1\%$ F1 increase) when tested on the target dataset’s text, compared to the much larger improvements ($\sim 8\%$ or more) obtained when the training and testing data were from the same dataset. This highlights a set challenges for future work in prompt optimization.

\subsubsection*{\textbf{Future Work}} We plan on exploring (a) cross-lingual prompt~\cite{DBLP:conf/emnlp/0001CWHC23} optimization, where we will apply automatic prompt optimization techniques to multilingual settings, and (b) interactive human-in-the-loop prompt optimization, i.e. combining automatic prompt optimization with human feedback loops. An interactive APO approach could allow a human analyst to guide the process by injecting domain knowledge or preferences (for example, selecting among automatically generated prompt candidates or refining them after a few iterations). Such approaches might fit well with agentic workflows that integrate tool-calling capabilities and decision making on when to get a human expert involved.

%Bibliography
\bibliographystyle{unsrt}  
\bibliography{references}

\end{document}